\documentclass[10pt,twocolumn,letterpaper]{article}

\usepackage{wacv}              

\usepackage[T1]{fontenc}
\usepackage{newtxtext,newtxmath}   
\usepackage{graphicx}
\usepackage{amsmath}
\usepackage{booktabs}
\usepackage{multirow}
\usepackage{algorithm}
\usepackage{algorithmic}
\usepackage{microtype}
\microtypesetup{expansion=false} 

\usepackage[pagebackref,breaklinks,colorlinks]{hyperref}

\usepackage[capitalize]{cleveref}
\crefname{section}{Sec.}{Secs.}
\Crefname{section}{Section}{Sections}
\Crefname{table}{Table}{Tables}
\crefname{table}{Tab.}{Tabs.}

\makeatletter
\let\origsubsubsection\subsubsection
\renewcommand{\subsubsection}{\@afterindenttrue\origsubsubsection}
\makeatother

\def\wacvPaperID{*****} 
\def\confName{WACV}
\def\confYear{2025}

\begin{document}

\title{Cross-Stage Attention Propagation for Efficient Semantic Segmentation}

\author{
Beoungwoo Kang \\
Vision Platform Team, Hyundai Mobis, South Korea\\
{\tt\small bwkang@mobis.com}
}

\maketitle

\begin{abstract}
Recent lightweight semantic segmentation methods have made significant progress by combining compact backbones with efficient decoder heads.
However, most multi-scale decoders compute attention independently at each feature scale, introducing substantial redundancy since the resulting attention distributions across scales are strongly correlated.
We propose \textbf{Cross-Stage Attention Propagation (CSAP)}, a decoder framework that computes attention at the deepest feature scale and propagates the resulting attention maps to shallower stages, bypassing query--key computation at those stages entirely.
This design preserves multi-scale contextual reasoning while substantially reducing the decoder's computational cost.
CSAP-Tiny achieves 42.9\% mIoU on ADE20K with only 5.5 GFLOPs, 80.5\% on Cityscapes with 21.5 GFLOPs, and 40.9\% on COCO-Stuff 164K with 5.5 GFLOPs, surpassing SegNeXt-Tiny by +1.8\% on ADE20K while requiring 16.8\% fewer floating-point operations.
\end{abstract}

\section{Introduction}
\label{sec:intro}

Semantic segmentation, the task of assigning a categorical label to every pixel in an image, is a fundamental problem in computer vision with wide-ranging applications in autonomous driving~\cite{cordts2016cityscapes} and robotic perception~\cite{garcia2017review}.
State-of-the-art segmentation architectures generally adopt an encoder--decoder paradigm, where a hierarchical encoder extracts multi-scale features and a decoder aggregates them into a dense prediction map.

\begin{figure}[t]
  \centering
  \includegraphics[width=\columnwidth]{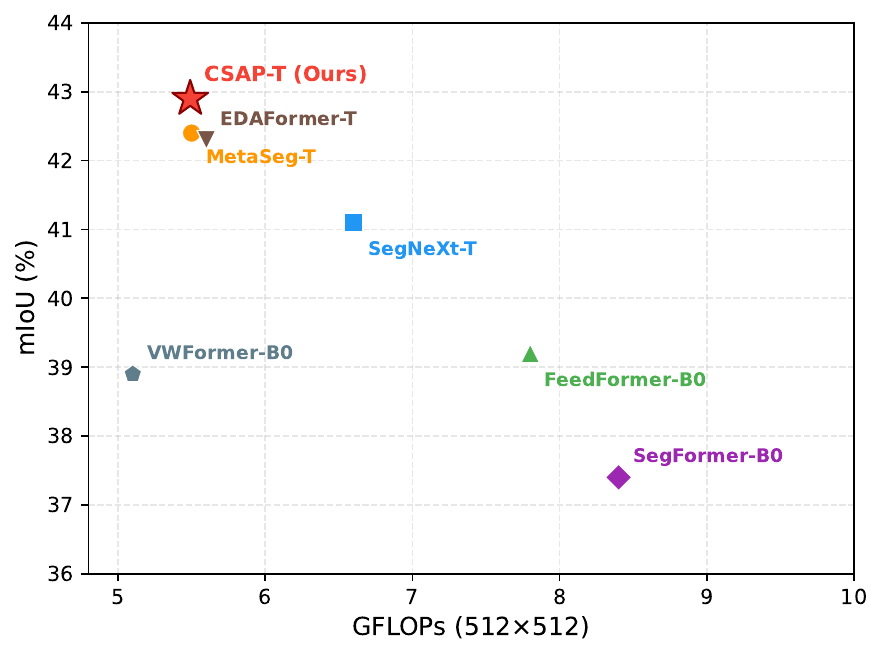}
  \caption{Performance comparison on ADE20K validation. Our CSAP-Tiny achieves 42.9\% mIoU with only 5.5 GFLOPs, outperforming SegNeXt-Tiny (41.1\%, 6.6 GFLOPs) with fewer computations and surpassing MetaSeg-T (42.4\%, 5.5 GFLOPs) and EDAFormer-T (42.3\%, 5.6 GFLOPs).}
  \label{fig:pareto}
\end{figure}

With the success of Vision Transformers~\cite{dosovitskiy2020image}, attention mechanisms have become a central component in dense prediction tasks, enabling effective capture of long-range spatial dependencies.
However, their computational cost grows quadratically with the number of tokens, making dense attention particularly expensive when applied across multiple decoder stages with high-resolution feature maps.
To address this limitation, various efficient designs have been proposed, including hierarchical attention~\cite{liu2021swin,wang2021pyramid}, mobile-oriented architectures~\cite{mehta2021mobilevit,chen2022mobile}, and lightweight convolutional backbones that achieve competitive representational capacity at reduced cost~\cite{guo2022segnext}.

A critical but overlooked source of inefficiency lies in how existing multi-scale decoders handle attention.
Most methods compute independent query--key dot products at each feature scale, despite the fact that the resulting attention distributions across different scales are strongly correlated.
Our analysis reveals that attention maps obtained at the deepest encoder stage closely resemble those computed independently at shallower stages, suggesting that a single attention computation at the deepest scale can effectively serve all stages.

Motivated by this observation, we propose \textbf{Cross-Stage Attention Propagation (CSAP)}, a decoder framework that computes attention only at the deepest feature scale and propagates the resulting attention maps to shallower stages.
At those stages, the decoder performs only value projections and directly applies the propagated attention weights, completely bypassing the expensive query--key dot-product computations that dominate the cost of conventional multi-scale decoders.
This eliminates the redundant per-stage attention computation while still enabling each stage to incorporate rich global contextual information derived from the deepest features.

We conduct extensive experiments on three challenging benchmarks to validate the effectiveness and efficiency of CSAP.
As shown in Figure~\ref{fig:pareto}, CSAP-Tiny achieves 42.9\% mIoU on ADE20K with only 5.5 GFLOPs, surpassing SegNeXt-Tiny by +1.8\% while requiring 16.8\% fewer floating-point operations.
It also outperforms MetaSeg-T~\cite{kang2024metaseg} and EDAFormer-T~\cite{yu2024edaformer} with marginally lower computational cost.
On Cityscapes and COCO-Stuff 164K, it records 80.5\% and 40.9\% mIoU respectively with 21.5 and 5.5 GFLOPs, consistently outperforming SegNeXt-Tiny across all three datasets.

Our main contributions are summarized as follows:
\begin{itemize}
  \item We propose CSAP, a lightweight decoder framework that eliminates redundant attention computation by propagating attention maps from the deepest feature scale to shallower stages, enabling efficient multi-scale contextual reasoning.
  \item We introduce an attention propagation mechanism that adapts the deepest-scale attention maps for use at shallower decoder stages, allowing those stages to bypass query--key computation entirely while preserving stage-specific spatial characteristics.
  \item Extensive experiments on ADE20K, Cityscapes, and COCO-Stuff 164K demonstrate that CSAP-Tiny achieves 42.9\%, 80.5\%, and 40.9\% mIoU respectively with only 5.5 GFLOPs on ADE20K, consistently surpassing existing lightweight segmentation methods at lower computational cost.
\end{itemize}

\section{Related Work}
\label{sec:related}

\subsection{Efficient Semantic Segmentation}

Semantic segmentation has progressed through several architectural generations.
Early fully convolutional networks (FCNs)~\cite{long2015fully} established the paradigm of dense prediction.
Subsequent works improved spatial detail recovery through encoder--decoder structures~\cite{ronneberger2015u,badrinarayanan2017segnet} and dilated convolutions~\cite{chen2017deeplab,yu2015multi}.

With the advent of attention-based models, SegFormer~\cite{xie2021segformer} demonstrated that a simple MLP decoder coupled with a hierarchical transformer backbone can yield strong performance.
SegNeXt~\cite{guo2022segnext} revisited the convolutional paradigm and showed that multi-scale convolutional attention can rival transformers while being more efficient.
FeedFormer~\cite{shim2023feedformer} proposed a feed-forward decoder that progressively refines multi-scale features.

On the efficiency front, TopFormer~\cite{zhang2022topformer} combined token pyramids from a lightweight backbone with a scale-aware injection module.
SeaFormer~\cite{wan2023seaformer} employed squeeze-enhanced axial attention to reduce the cost of long-range modeling.
RTFormer~\cite{wang2022rtformer} integrated attention and convolution in a dual-branch architecture for real-time inference.
MetaSeg~\cite{kang2024metaseg} proposed a MetaFormer-based decoder that captures global contexts through efficient token mixing.
VWFormer~\cite{yan2024vwformer} introduced varying window attention to capture multi-scale representations adaptively, and EDAFormer~\cite{yu2024edaformer} reduced spatial redundancy at inference time by selectively discarding tokens.

Despite these advances, most methods compute attention independently at each feature scale, leading to redundant computations.
Our work differs by sharing attention maps across scales, reducing cost while maintaining multi-scale reasoning.

\subsection{Hierarchical Multi-Scale Representations}

Extracting features at multiple spatial scales is central to dense prediction.
Convolutional backbones such as ResNet~\cite{he2016deep} naturally produce a hierarchy of feature maps with decreasing resolution and increasing semantics.
FPN~\cite{lin2017feature} enriches this hierarchy with a top-down pathway, while HRNet~\cite{sun2019deep} maintains parallel multi-resolution streams and exchanges information across them.
Hierarchical transformers such as PVT~\cite{wang2021pyramid} and Swin Transformer~\cite{liu2021swin} adopt a similar staged design, and SegNeXt~\cite{guo2022segnext} captures multi-scale contexts through multi-branch depth-wise convolutions in the MSCAN encoder.
DeepLabv3+~\cite{chen2018encoder} combines atrous spatial pyramid pooling with an encoder--decoder structure for effective multi-scale reasoning.
ConvNeXt~\cite{liu2022convnet} further showed that modernized convolutional designs can match hierarchical transformers on dense prediction benchmarks.

While these works focus on \emph{how to extract} hierarchical features, most decoders simply concatenate or sum them after independent per-scale processing.
Our CSAP decoder addresses this gap by propagating attention from the coarsest scale to finer ones, reusing high-level patterns instead of recomputing them.

\subsection{Attention Sharing and Reuse}

While attention mechanisms have been extensively studied in vision models, the idea of sharing attention maps across different feature scales remains underexplored.

Several prior works have explored reducing the spatial dimension of keys and values to lower the cost of attention.
PVT~\cite{wang2021pyramid} employs strided convolutions to spatially reduce keys and values, while PVTv2~\cite{wang2022pvtv2} replaces this with average pooling for improved efficiency.
However, these approaches still compute query--key products at every scale independently.
Our approach goes further by computing attention once and propagating the resulting maps to other scales, eliminating the need for any query--key computation at those scales.
This strategy is complementary to efficient attention designs such as linear attention~\cite{kitaev2020reformer} or window attention~\cite{liu2021swin}, and can be combined with them for further efficiency gains.

\section{Method}
\label{sec:method}

\begin{figure*}[t]
  \centering
  \includegraphics[width=\textwidth]{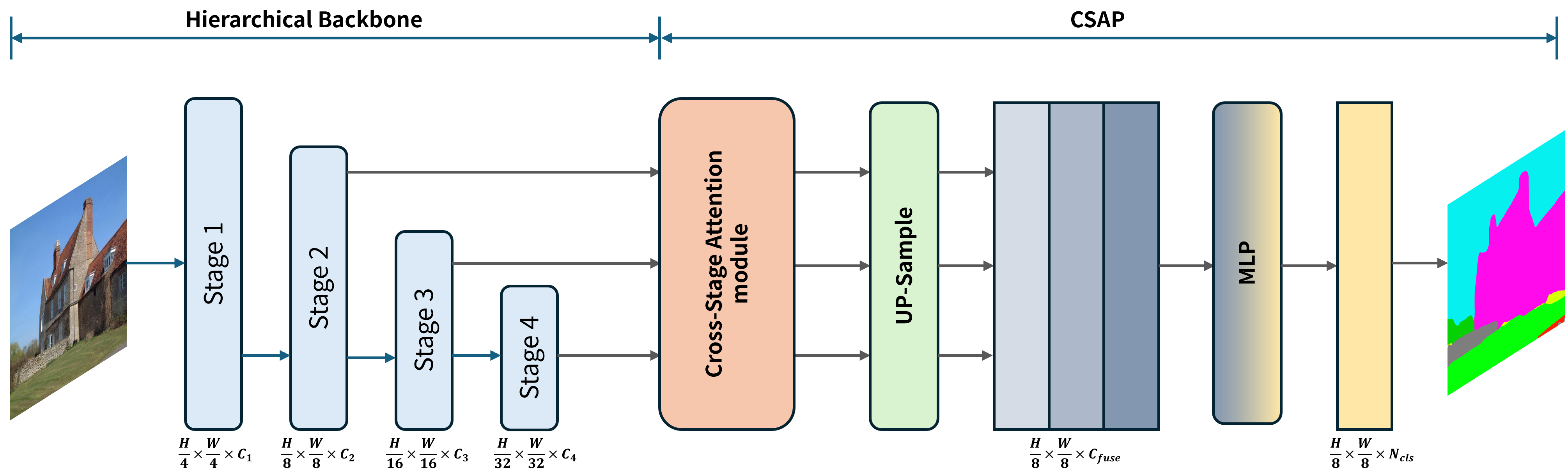}
  \caption{Overall architecture of the proposed CSAP framework. The hierarchical backbone extracts multi-scale features $\{C_2, C_3, C_4\}$ at resolutions $\{1/8, 1/16, 1/32\}$. Cross-attention is computed at stage~4 with spatially pooled key--value tokens, producing attention maps $A_4$. These maps are propagated to stages~2 and~3, where only value projections are computed while Q--K computation is bypassed.}
  \label{fig:architecture}
\end{figure*}

\begin{figure}[t]
  \centering
  \includegraphics[width=\columnwidth]{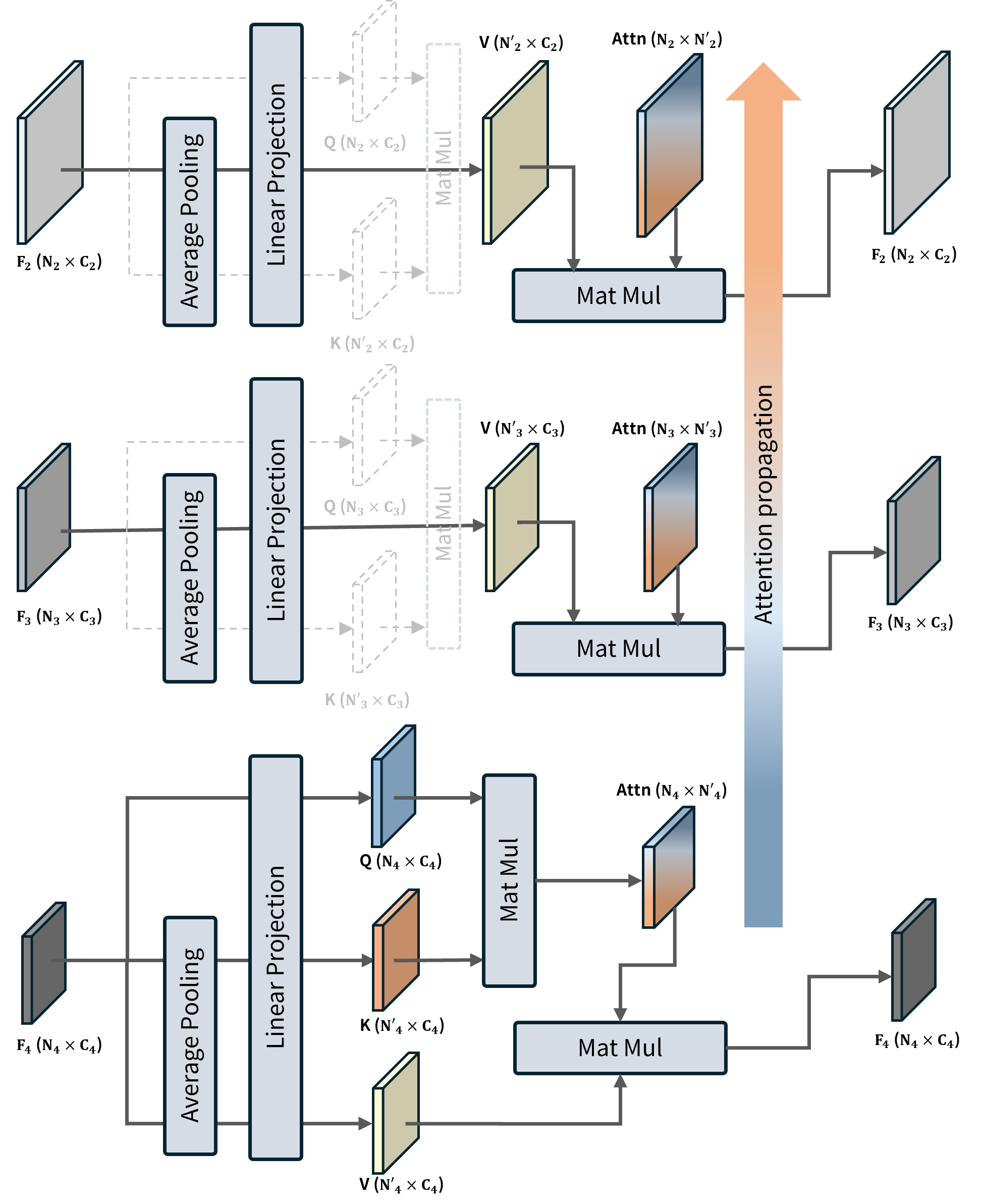}
  \caption{Detailed structure of the Cross-Stage Attention module. At stage~4, pooled cross-attention computes Q from the original tokens and K, V from spatially pooled context tokens, producing multi-head attention maps $A_4$. The attention maps are reshaped, adaptively pooled to a fixed $8 \times 8$ spatial size, and transformed to produce stage-specific attention weights $A_{4 \rightarrow 2}$ and $A_{4 \rightarrow 3}$. At stages~2 and~3, only value projections are computed and combined with the propagated attention weights.}
  \label{fig:module}
\end{figure}

We present Cross-Stage Attention Propagation (CSAP), a lightweight decoder framework for efficient semantic segmentation.
Figure~\ref{fig:architecture} provides an overview of the proposed architecture, and Figure~\ref{fig:module} illustrates the detailed structure of the Cross-Stage Attention module.

\subsection{Overall Architecture}
\label{sec:overall}

\subsubsection{Hierarchical Convolutional Encoder}
\label{sec:encoder}

We adopt the MSCAN (Multi-Scale Convolutional Attention Network)~\cite{guo2022segnext} as the feature encoder.
MSCAN is a purely convolutional backbone that captures multi-scale contextual information efficiently, producing a hierarchy of feature maps at four spatial scales.
Given an input image $I \in \mathbb{R}^{H \times W \times 3}$, the encoder produces feature maps $\{C_1, C_2, C_3, C_4\}$ at resolutions $\{1/4, 1/8, 1/16, 1/32\}$ of the input, with channel dimensions $\{32, 64, 160, 256\}$ for the Tiny variant.
This convolutional backbone provides an efficient foundation with strong local feature extraction, onto which our attention-based decoder adds global contextual reasoning.

\subsubsection{Cross-Stage Attention Decoder}
\label{sec:decoder}

Our decoder operates on features from stages 2 through 4, which span a sufficient range of spatial detail and semantic abstraction.
At stage~4, a pooled cross-attention block processes $C_4$ to produce refined features $\hat{C}_4$ and attention maps $A_4$.
The attention maps are then reshaped, pooled to a fixed spatial size, and transformed into stage-specific attention weights $A_{4 \rightarrow 2}$ and $A_{4 \rightarrow 3}$.
At stages~2 and~3, each block computes only value tokens and applies the propagated attention maps, bypassing Q--K computation entirely.
After processing, the refined features from all three stages are resized to the stage-2 resolution ($1/8$ of input), concatenated along the channel dimension, and fused through a $1\!\times\!1$ convolution followed by a classifier head.
The final prediction $\hat{Y} \in \mathbb{R}^{B \times K \times H_2 \times W_2}$ is bilinearly upsampled to the original resolution during inference.

\subsection{Cross-Stage Attention Propagation}
\label{sec:csap}

\subsubsection{Pooled Cross-Attention}
\label{sec:cross_attention}

To capture global context while keeping computation tractable, we employ a pooled cross-attention mechanism at stage~4, where the feature map has the smallest spatial resolution of $16 \times 16$ for a $512 \times 512$ input and the richest semantic abstraction.
Given a feature map $C \in \mathbb{R}^{B \times N' \times D}$ where $N' = H' \times W'$ is the number of spatial tokens, the standard self-attention requires $\mathcal{O}(N'^2)$ complexity.
We reduce this by generating compact context tokens via spatial average pooling.

The feature map is first reshaped to its spatial layout and downsampled by a factor of $r$:
\begin{equation}
  \bar{C} = \text{AvgPool}_{r}\!\left(\text{Reshape}(C)\right) \in \mathbb{R}^{B \times D \times \frac{H'}{r} \times \frac{W'}{r}}
\end{equation}
A $1\times1$ convolution followed by layer normalization and GELU is applied:
\begin{equation}
  \bar{C}' = \text{GELU}\!\left(\text{LN}\!\left(\text{Conv}_{1\times1}(\bar{C})\right)\right)
\end{equation}
Query, key, and value are computed as:
\begin{align}
  Q &= C \cdot W_Q \in \mathbb{R}^{B \times N' \times d}, \\
  K &= \bar{C}' \cdot W_K \in \mathbb{R}^{B \times M \times d}, \\
  V &= \bar{C}' \cdot W_V \in \mathbb{R}^{B \times M \times d},
\end{align}
where $M = \frac{H'}{r} \times \frac{W'}{r}$ and $d$ is the projected dimension.
Multi-head attention is computed as:
\begin{equation}
  A = \text{softmax}\!\left(\frac{Q K^\top}{\sqrt{d_h}}\right)\!, \quad \hat{C} = A V
  \label{eq:attention}
\end{equation}
where $d_h = d / n_\text{heads}$.

Each cross-attention operation is followed by a feed-forward network (FFN) with depth-wise convolution~\cite{xie2021segformer} and residual connections:
\begin{equation}
  x' = x + \text{Attn}(\text{LN}(x)), \;\; x'' = x' + \text{FFN}(\text{LN}(x'))
\end{equation}

\subsubsection{Attention Propagation and Refinement}
\label{sec:propagation}

The central innovation of our decoder is the attention propagation mechanism.
Rather than computing Q--K attention independently at each stage, we compute it once at stage~4 and propagate the resulting attention maps to stages~2 and~3.

\paragraph{Attention Reshaping and Projection.}
At stage~4, the cross-attention block produces both refined features $\hat{C}_4$ and the multi-head attention weight matrix $A_4 \in \mathbb{R}^{B \times n_h \times N'_4 \times M_4}$, where $n_h$ is the number of heads, $N'_4 = H_4 \times W_4$, and $M_4$ is the number of pooled context tokens.
The attention maps are reorganized to expose their spatial structure.
$A_4$ is permuted and reshaped into a 2D spatial form, then adaptively pooled to a fixed spatial size $s \times s$:
\begin{equation}
  \tilde{A}_4 = \text{AdaptiveAvgPool}_{s \times s}\!\left(\text{Reshape}(A_4)\right)
\end{equation}
where $s = 8$.
This produces a compact attention representation that is flattened and reshaped into multi-head format: $\tilde{A}_4 \in \mathbb{R}^{B \times n_h \times L \times s^2}$.
Two stage-specific transformations $\text{Proj}_{\rightarrow 2}$ and $\text{Proj}_{\rightarrow 3}$ convert the shared attention into stage-specific weights:
\begin{align}
  A_{4 \rightarrow 2} &= \text{Proj}_{\rightarrow 2}(\tilde{A}_4), \\
  A_{4 \rightarrow 3} &= \text{Proj}_{\rightarrow 3}(\tilde{A}_4),
\end{align}
which adapt the deepest-scale attention patterns to the spatial characteristics of stages~2 and~3.

\paragraph{Value-Only Refinement.}
At stages~2 and~3, we bypass the query--key computation entirely and use the propagated attention maps directly.
Given stage features $C_k$ for $k \in \{2, 3\}$ and propagated attention $A_{4 \rightarrow k}$, value tokens are first obtained by applying a value embedding to the stage features:
\begin{equation}
  V_k = C_k \cdot W_V
\end{equation}
The refined output $\hat{C}_k$, which denotes the attention-weighted feature representation at stage $k$, is computed as:
\begin{equation}
  \hat{C}_k = A_{4 \rightarrow k} \cdot V_k
\end{equation}
This eliminates the query--key computation at each stage, replacing it with a simple matrix multiplication using pre-computed attention weights.
The refined features are then processed through an FFN with residual connections.

\section{Experiments}
\label{sec:experiments}

\begin{table*}[!t]
  \centering
  \small
  \caption{Comparison with state-of-the-art lightweight semantic segmentation methods on ADE20K, Cityscapes, and COCO-Stuff. Only tiny/small-scale models are compared. GFLOPs are measured at the resolution used for each dataset.}
  \label{tab:main}
  \resizebox{\textwidth}{!}{
  \begin{tabular}{l|c|cc|cc|cc}
    \toprule
    \multirow{2}{*}{Method} & \multirow{2}{*}{Params (M) $\downarrow$} & \multicolumn{2}{c|}{ADE20K} & \multicolumn{2}{c|}{Cityscapes} & \multicolumn{2}{c}{COCO-Stuff} \\
    \cmidrule(lr){3-4} \cmidrule(lr){5-6} \cmidrule(lr){7-8}
     & & GFLOPs $\downarrow$ & mIoU(\%) $\uparrow$ & GFLOPs $\downarrow$ & mIoU(\%) $\uparrow$ & GFLOPs $\downarrow$ & mIoU(\%) $\uparrow$ \\
    \midrule
    TopFormer-B~\cite{zhang2022topformer} & 5.1 & 1.8 & 39.2 & 5.7 & 71.2 & -- & -- \\
    SegFormer-B0~\cite{xie2021segformer} & 3.8 & 8.4 & 37.4 & 125.5 & 76.2 & 8.4 & 35.6 \\
    FeedFormer-B0~\cite{shim2023feedformer} & 4.5 & 7.8 & 39.2 & 107.4 & 77.9 & -- & -- \\
    VWFormer-B0~\cite{yan2024vwformer} & 3.7 & 5.1 & 38.9 & -- & 77.2 & 5.1 & 36.2 \\
    MetaSeg-T~\cite{kang2024metaseg} & 4.7 & 5.5 & 42.4 & 71.7 & 80.1 & 5.5 & 39.7 \\
    EDAFormer-T~\cite{yu2024edaformer} & 4.9 & 5.6 & 42.3 & 151.7 & 78.7 & 5.6 & 40.3 \\
    SegNeXt-T~\cite{guo2022segnext} & 4.3 & 6.6 & 41.1 & 56.0 & 79.8 & 6.6 & 38.8 \\
    \midrule
    \textbf{CSAP-T (Ours)} & \textbf{4.8} & \textbf{5.5} & \textbf{42.9} & \textbf{21.5} & \textbf{80.5} & \textbf{5.5} & \textbf{40.9} \\
    \bottomrule
  \end{tabular}
  }
\end{table*}

\subsection{Experimental Settings}
\label{sec:settings}

\paragraph{Datasets.}
We evaluate CSAP on three standard semantic segmentation benchmarks.
ADE20K~\cite{zhou2017scene} is a large-scale scene parsing benchmark with 150 semantic classes, split into 20,210 training and 2,000 validation images.
Cityscapes~\cite{cordts2016cityscapes} focuses on urban street-level understanding with 19 object categories, consisting of 2,975 finely annotated training images and 500 validation images at a native resolution of $2048 \times 1024$.
COCO-Stuff 164K~\cite{caesar2018coco} provides exhaustive pixel-level annotations for 171 categories (80 object and 91 stuff classes) across 118,287 training and 5,000 validation images.

\paragraph{Implementation Details.}
All experiments are built on MMSegmentation~\cite{mmseg2020} and conducted on NVIDIA RTX 3090 GPUs.
Optimization is performed using AdamW~\cite{loshchilov2018decoupled} with an initial learning rate of $6 \times 10^{-5}$ and a weight decay of $0.01$.
The learning rate follows a polynomial decay schedule (power $= 1.0$) with a linear warm-up over the first 1,500 iterations.
For ADE20K and COCO-Stuff 164K, we set the training crop to $512 \times 512$ with a batch size of 16 and train for 160K iterations.
For Cityscapes, the crop size is $1024 \times 1024$ with a batch size of 8 over 160K iterations.
Standard data augmentation strategies are applied, including random resizing with a scale factor in $[0.5, 2.0]$, random horizontal flipping, and photometric distortion.
The CSAP decoder is configured with 4 attention heads, a projected dimension of $d=128$, a pool ratio of $r=2$, an FFN expansion ratio of 4, and an adaptive pooling target size of $s=8$ for attention propagation.
Performance is measured using the mean Intersection-over-Union (mIoU) metric on each validation set.

\subsection{Comparison with State-of-the-Art}
\label{sec:main_results}

Table~\ref{tab:main} summarizes the quantitative comparison between CSAP and recent lightweight segmentation methods across three benchmarks.
On ADE20K, CSAP-T achieves 42.9\% mIoU with 5.5 GFLOPs.
This represents a +1.8\% gain over SegNeXt-T while requiring 16.8\% fewer floating-point operations.
CSAP-T also surpasses MetaSeg-T by +0.5\% and EDAFormer-T by +0.6\%, both with marginally lower computational cost.
On Cityscapes, our method records 80.5\% mIoU with only 21.5 GFLOPs, exceeding SegNeXt-T by +0.7\%.
On COCO-Stuff 164K, CSAP-T reaches 40.9\% mIoU, outperforming SegNeXt-T by +2.1\% and EDAFormer-T by +0.6\%.
Across all three datasets, CSAP-T achieves the highest mIoU among comparable methods, confirming that attention propagation is a highly effective strategy for building efficient yet accurate decoders.

\subsection{Ablation Studies}
\label{sec:ablation}

We perform ablation experiments on the ADE20K validation set to examine the contribution of each design choice in CSAP.
All variants share the same MSCAN-T backbone and training recipe.

\subsubsection{Effect of Decoder Design}
\label{sec:ablation_decoder}

\begin{table}[t]
  \centering
  \small
  \caption{Effect of attention propagation on ADE20K. Standard computes full Q, K, V attention at all three stages independently. CSAP computes Q, K, V only at stage~4 and propagates the attention maps to stages~2 and~3. Both use the same MSCAN-T backbone.}
  \label{tab:ablation}
  \setlength{\tabcolsep}{5pt}
  \begin{tabular}{l|c|c|c}
    \toprule
    Decoder Design & Params $\downarrow$ & GFLOPs $\downarrow$ & mIoU(\%) $\uparrow$ \\
    \midrule
    Hamburger~\cite{guo2022segnext} & 4.3M & 6.6 & 41.1 \\
    Standard Self-Attn~\cite{xie2021segformer} & 5.0M & 5.9 & 43.1 \\
    \textbf{CSAP (Ours)} & \textbf{4.8M} & \textbf{5.5} & \textbf{42.9} \\
    \bottomrule
  \end{tabular}
\end{table}

Table~\ref{tab:ablation} contrasts three decoder architectures mounted on the same backbone.
The standard self-attention variant independently applies full Q, K, V projections and dot-product attention at every decoder stage.
CSAP replaces the per-stage Q--K computation at stages~2 and~3 with propagated attention maps generated at stage~4.
Although the standard baseline yields the highest absolute mIoU at 43.1\%, CSAP trails by only 0.2\% while consuming 6.5\% fewer GFLOPs and 5.2\% fewer parameters.
Both attention-based decoders substantially outperform the Hamburger decoder used in SegNeXt at 41.1\%, with CSAP delivering a +1.8\% improvement at 16.8\% lower computational cost.

\subsubsection{Effect of Propagation Source Stage}
\label{sec:ablation_stage}

\begin{table}[t]
  \centering
  \small
  \caption{Ablation on the propagation source stage. Full Q, K, V attention is computed at the source stage, and the resulting attention maps are propagated to the remaining two stages. All models use the MSCAN-T backbone on ADE20K.}
  \label{tab:stage_ablation}
  \setlength{\tabcolsep}{8pt}
  \begin{tabular}{c|c|c|c}
    \toprule
    Source Stage & Params $\downarrow$ & GFLOPs $\downarrow$ & mIoU(\%) $\uparrow$ \\
    \midrule
    Stage 2 ($1/8$) & 5.3M & 6.1 & 42.4 \\
    Stage 3 ($1/16$) & 5.1M & 5.7 & 42.4 \\
    \textbf{Stage 4 ($1/32$)} & \textbf{4.8M} & \textbf{5.5} & \textbf{43.0} \\
    \bottomrule
  \end{tabular}
\end{table}

Table~\ref{tab:stage_ablation} examines which encoder stage should serve as the source for attention generation.
We test three configurations: propagating from stage~2 at $1/8$ resolution with $64 \times 64$ spatial size, stage~3 at $1/16$ with $32 \times 32$, and stage~4 at $1/32$ with $16 \times 16$.
Stage~4 obtains the best accuracy at 43.0\% while requiring the fewest parameters and lowest computational budget at 5.5 GFLOPs.
This outcome aligns with the intuition that the deepest encoder features, being the most semantically abstracted, produce attention maps that generalize well when transferred to shallower stages.
In contrast, stage~2 incurs the highest cost at 6.1 GFLOPs due to its larger spatial dimensions, yet produces the weakest accuracy at 42.4\%, indicating that spatially detailed but semantically shallow features yield less transferable relational patterns.

\subsubsection{Design Analysis}
\label{sec:ablation_analysis}

The attention propagation mechanism introduces minimal overhead.
Each of the two projection functions $\text{Proj}$ operates on $s^2 \times s^2 = 64 \times 64$ entries, contributing only 8K parameters in total.
In terms of computation, the Q--K dot product at stage~4 operates over a $16 \times 16$ spatial resolution, and stages~2 and~3 bypass Q--K computation entirely, reducing the decoder's matrix-multiplication FLOPs from 0.18G to 0.01G.

\subsubsection{Attention Map Visualization}
\label{sec:attn_vis}

\begin{figure}[t]
  \centering
  \includegraphics[width=\columnwidth]{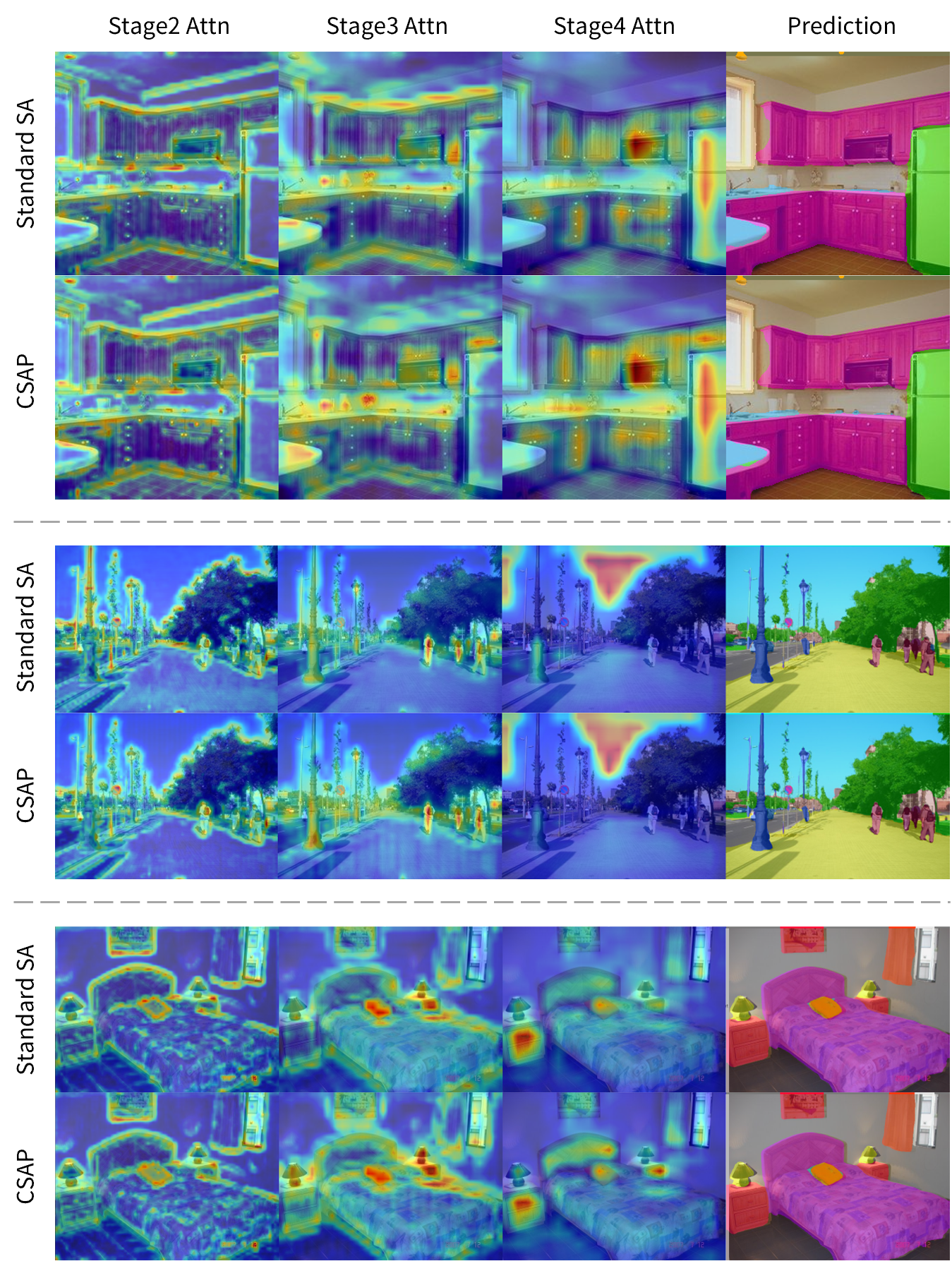}
  \caption{Attention map visualization comparing standard self-attention (top) and cross-stage attention propagation (bottom). Despite computing Q--K attention only at stage~4, CSAP produces attention patterns highly similar to those of the standard baseline across all three stages, yielding nearly identical predictions. This validates that per-stage Q--K computation is largely redundant.}
  \label{fig:attention}
\end{figure}

To qualitatively verify the effectiveness of attention propagation, we visualize the multi-head attention maps at each decoder stage in Figure~\ref{fig:attention}.
In the standard self-attention variant, where Q--K products are independently computed at every stage, the resulting spatial attention distributions across stages are largely overlapping, confirming the redundancy motivating our approach.
CSAP's propagated attention maps, derived from a single computation at stage~4, exhibit highly similar spatial structure to the per-stage baseline.
This consistency indicates that the deepest-scale attention prior effectively encodes the relational information needed at all scales, while the stage-specific projections fine-tune it for the respective spatial resolutions.

\subsection{Qualitative Results}

\begin{figure}[t]
  \centering
  \includegraphics[width=\columnwidth]{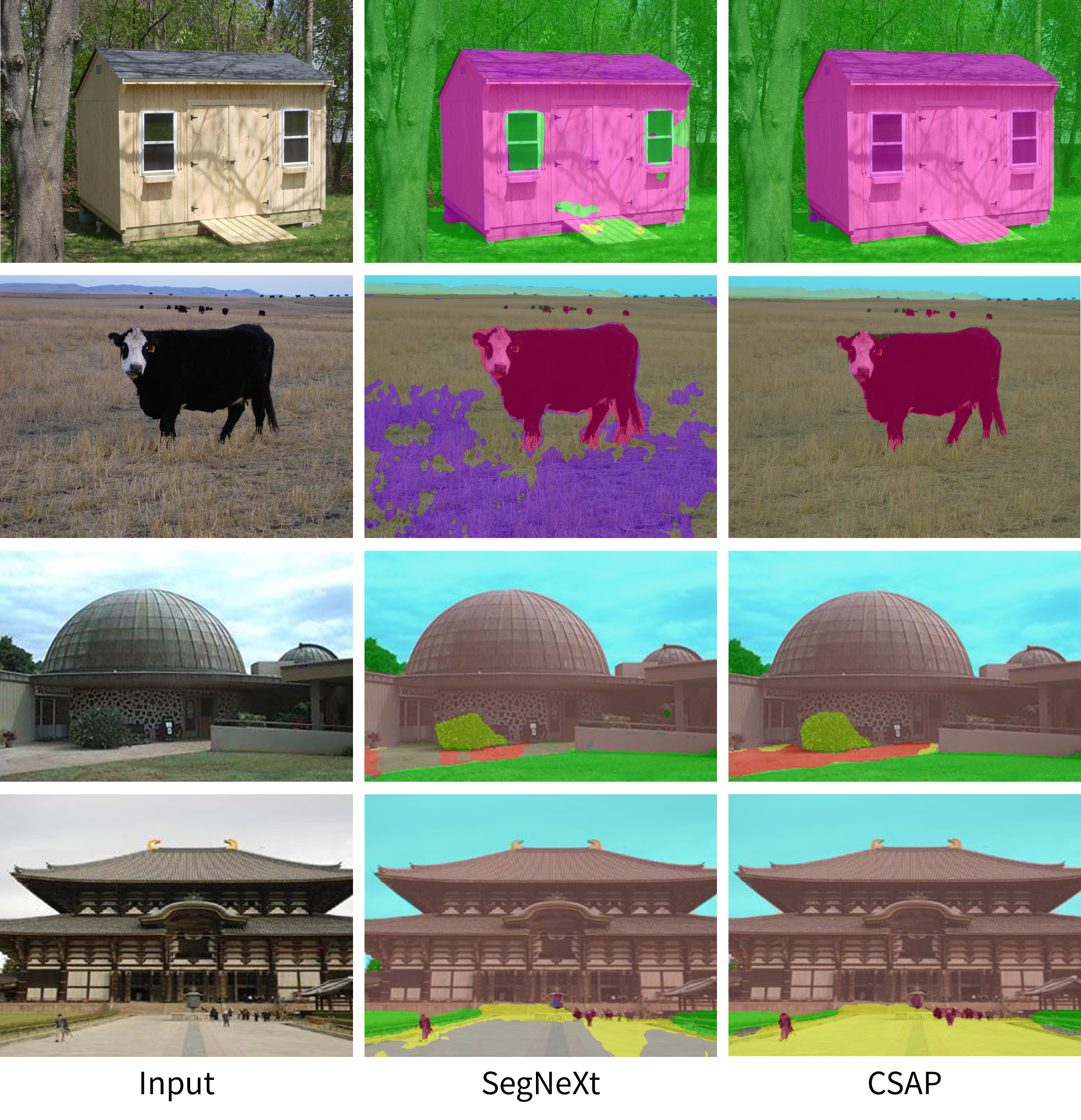}
  \caption{Qualitative segmentation results on ADE20K validation. From left to right: Input image, SegNeXt-T prediction, and CSAP-T prediction. Our CSAP-T produces more coherent predictions on global regions.}
  \label{fig:qualitative_ade}
\end{figure}

\begin{figure}[t]
  \centering
  \includegraphics[width=\columnwidth]{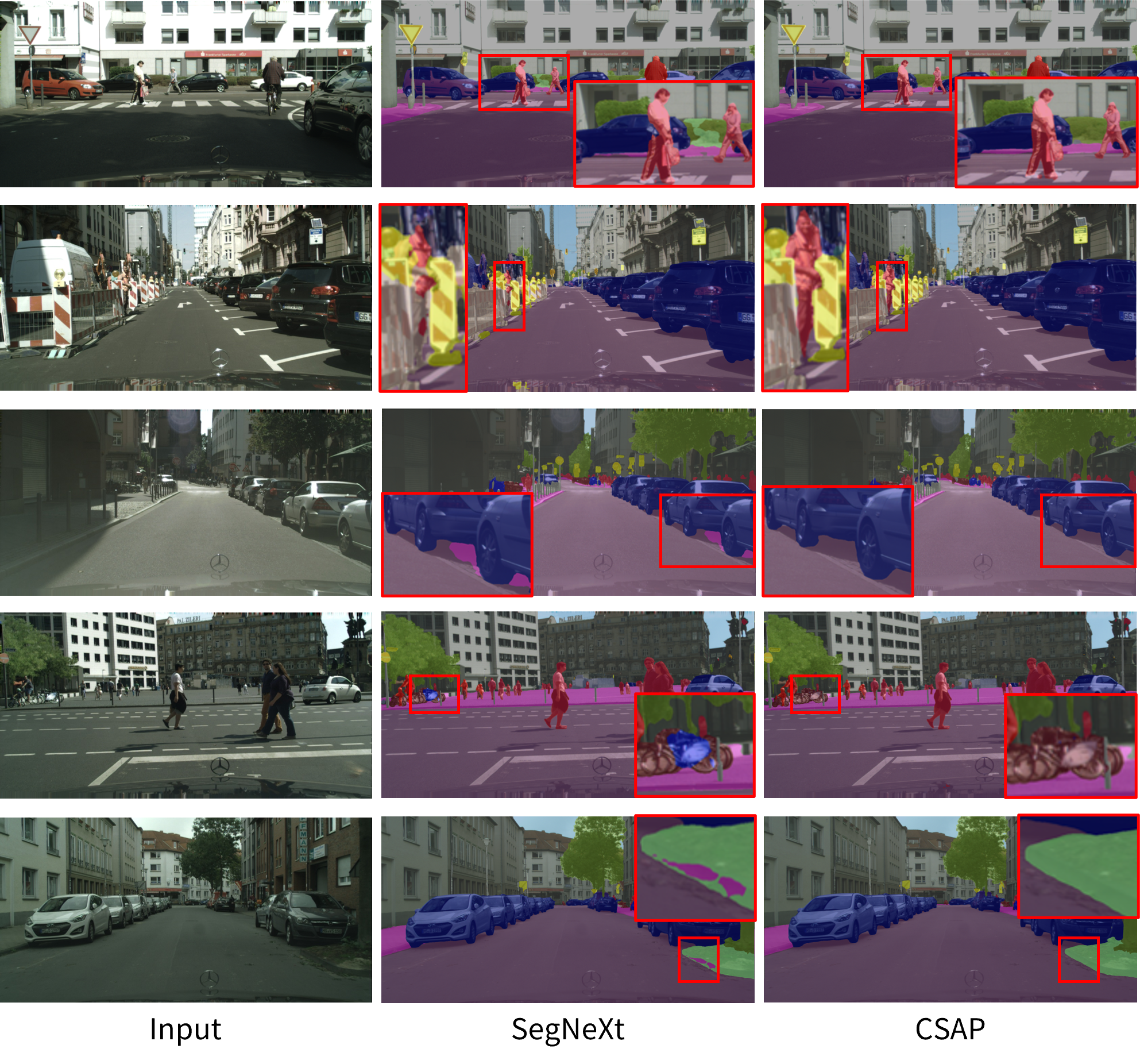}
  \caption{Qualitative segmentation results on Cityscapes validation. From left to right: Input image, SegNeXt-T prediction, and CSAP-T prediction.}
  \label{fig:qualitative_city}
\end{figure}

Figure~\ref{fig:qualitative_ade} and Figure~\ref{fig:qualitative_city} present qualitative comparisons on ADE20K and Cityscapes, where CSAP-T visibly outperforms SegNeXt-T on large homogeneous regions and small ambiguous objects.

\paragraph{ADE20K.}
On ADE20K, CSAP-T demonstrates stronger segmentation of large-scale global regions including houses, grasslands, and ground areas.
These broad regions are predicted with notably greater spatial coherence compared to SegNeXt-T, which tends to fragment or misclassify parts of these areas.
For example, ground regions are segmented much more completely, confirming that the propagated attention maps effectively capture global contextual relationships across the scene.

\paragraph{Cityscapes.}
On Cityscapes, CSAP-T shows clear improvements on challenging elements: distant vegetation that blends with the background is better delineated, half-occluded pedestrians are more reliably recognized, ground surfaces beneath vehicles are correctly classified, and distant hard-to-distinguish objects such as bicycles are segmented more accurately.

These results suggest that attention propagation enables the decoder to maintain discriminative power even for ambiguous regions that require both local detail and global context.
These improvements are consistent with the attention map analysis in Section~\ref{sec:attn_vis}: the propagated attention distributions closely match those of the full per-stage computation, confirming that cross-stage sharing preserves the spatial specificity needed for precise boundary delineation.
Overall, these qualitative observations corroborate the quantitative results in Table~\ref{tab:main} and validate that attention propagation from the deepest scale transfers rich contextual information across all decoder stages.

\section{Conclusion}
\label{sec:conclusion}

We presented Cross-Stage Attention Propagation (CSAP), a decoder framework that computes attention at the deepest feature scale and propagates the resulting attention maps to shallower stages, eliminating redundant query--key computations across the decoder hierarchy.
Built upon the observation that attention distributions across multi-scale decoder stages are strongly correlated, CSAP enables efficient multi-scale contextual reasoning by requiring only value projections at the shallower stages.
CSAP-T achieves 42.9\% mIoU on ADE20K with only 5.5 GFLOPs, 80.5\% on Cityscapes with 21.5 GFLOPs, and 40.9\% on COCO-Stuff 164K with 5.5 GFLOPs, consistently outperforming existing lightweight segmentation methods at lower computational cost.
Our ablation studies confirm that the deepest encoder stage produces the most transferable attention maps, and that propagation introduces negligible overhead while preserving segmentation quality within 0.2\% of the full per-stage attention baseline.
These results demonstrate that attention reuse across feature hierarchies is a highly effective strategy for building efficient yet accurate decoders.
As attention propagation is orthogonal to backbone design and training recipe, it can serve as a general plug-in component for future hierarchical segmentation architectures.

\subsection{Future Work}

Since the proposed attention propagation mechanism is architecture-agnostic, it can be readily integrated into other hierarchical encoder--decoder frameworks.
We plan to scale CSAP to larger model variants and evaluate its generality across different backbone families.
An important direction is investigating whether attention maps from intermediate scales can serve as complementary propagation sources, potentially capturing different levels of semantic granularity.
Additionally, extending the propagation strategy to other dense prediction tasks such as object detection, depth estimation, and panoptic segmentation is a promising avenue for future research.

\clearpage
{\small
\bibliographystyle{ieee_fullname}
\bibliography{references}
}

\appendix
\section{The Use of Large Language Models (LLMs)}
\label{sec:llm}

Throughout the course of writing this paper, large language models were utilized under careful supervision and in a strictly auxiliary role.
Their contribution was limited to English proofreading and grammatical refinement of the manuscript text.
They were not used to compose full sentences or paragraphs, nor to produce novel ideas, methodologies, or experimental results.
All research contributions presented in this work---including the proposed architecture, experimental design, and analysis---originate solely from the author.

\section*{Acknowledgements}

I thank Hyunwoo Yu, Yubin Cho, Seunghun Moon, Jouwon Song, Jincheol Yang, and Matti Zinke for the opportunity to work together.
I also thank Prof.\ Suk-Ju Kang for his support.

\end{document}